\documentclass[runningheads,a4paper]{llncs}

\usepackage{xargs}                    
\usepackage[pdftex,dvipsnames]{xcolor} 
\usepackage[colorinlistoftodos,prependcaption,textsize=tiny]{todonotes}
\newcommandx{\unsure}[2][1=]{\todo[linecolor=red,backgroundcolor=red!25,bordercolor=red,#1]{#2}}
\newcommandx{\change}[2][1=]{\todo[linecolor=blue,backgroundcolor=blue!25,bordercolor=blue,#1]{#2}}
\newcommandx{\info}[2][1=]{\todo[linecolor=OliveGreen,backgroundcolor=OliveGreen!25,bordercolor=OliveGreen,#1]{#2}}

\usepackage{graphicx}
\usepackage{sidecap}
\sidecaptionvpos{figure}{c}
\usepackage{array}

\usepackage{amssymb,amsmath}

\usepackage[T1]{fontenc}
\usepackage{cite}
\usepackage{inconsolata} 

\usepackage{tikz-cd}
\usepackage{mathtools}
\usepackage{amsfonts}

\newtheorem{mydef}{Definition}
\newtheorem{myprob}{Problem}

\newtheorem{myth}{Theorem}
\newtheorem{est}{Estimator}

\newcommand{\bM}{{\sf{M}}}
\newcommand{\ebM}{{\hat{\sf{M}}}}
\newcommand{\bm}{{\sf{m}}}

\newcommand{\cvar}{{\sf{X}}}

\newcommand{\pfpar}{\bth_{\sf{f}}}
\newcommand{\pmpar}{\bth_{\sf{m}}}
\newcommand{\pMpar}{\bth_{\sf{M}}}
\newcommand{\rarg}[1]{{\bth_{\bm}^{(#1)}}}

\newcommand{\tstat}{\eta} 
\newcommand{\stat}{{\hat{\eta}}} 

\newcommand{\tmap}{{{\cal M}_\tstat}}
\newcommand{\map}{{{\cal M}_\stat}}

\newcommand{\bth}{\boldsymbol{\theta}}
\newcommand{\xx}{\boldsymbol{x}}

\newcommand{\R}{\mathbb{R}}

\newcommand{\tmapi}[1]{{\cal M}_{\tstat,#1}}

\newcommand{\var}{{\sf Var}}

\newcommand{\MATLAB}{{\sf MATLAB}}

\newcommand{\QSSA}{{\sf QSSA}}
\newcommand{\deriv}[1]{\dot{{#1}}} 

\newcommand{\E}{{\sf{E}}} 
\newcommand{\ES}{{{\sf{ES}}} }
\newcommand{\SU}{{{\sf{S}}}} 
\newcommand{\PR}{{{\sf{PR}}}} 

\newcommand{\COND}{{{\bf  C}}} 

\newcommand{\PTN}{{\sf PTN}}
\newcommand{\GE}{{\sf{G}}}
\newcommand{\GEI}{{\sf{G}_{\text{\sf in}}}}
\newcommand{\GEA}{{\sf{G}_{\text{\sf act}}}}
\newcommand{\RNA}{{\sf{mRNA}}}

\begin{document}

\mainmatter  

\title{Matching models across abstraction levels with Gaussian Processes\thanks{GC and GS gratefully acknowledge support from the European Research Council under grant MLCS306999. LB acknowledges partial support from the EU project QUANTICOL, 600708, and by FRA-UniTS. We thank Dimitris Milios for useful discussions and for providing us with the \MATLAB{} for heteroscedastic regression.}}

\titlerunning{Matching models across abstraction levels with Gaussian Processes}

\author{Giulio Caravagna\inst{1}%
\thanks{Corresponding author, {\sf giulio.caravagna@ed.ac.uk}.}%
\and Luca Bortolussi\inst{2,3}\and Guido Sanguinetti\inst{1}}

\authorrunning{Giulio Caravagna, Luca Bortolussi and Guido Sanguinetti}
\institute{School of Informatics, University of Edinburgh, UK.
\and
 DMG, University of Trieste, Trieste, Italy;\quad  ISTI-CNR, Pisa, Italy.
 \and
 MOSI, Dept. of Informatics, Saarland University, Saarb\"ucken, Germany.}

\toctitle{Matching models across abstraction levels with Gaussian Processes}
\tocauthor{Giulio Caravagna {\em et al.}}

%

\maketitle

\begin{abstract}
Biological systems are often modelled at different levels of abstraction depending on the particular aims/resources of a study. Such different models often provide qualitatively concordant predictions  over specific parametrisations, but it is generally unclear whether model predictions are quantitatively in agreement, and whether such agreement holds for different parametrisations. Here we present a generally applicable statistical machine learning methodology to automatically reconcile the predictions of different models across abstraction levels. Our approach is based on defining a correction map, a random function which modifies the output of a model in order to match the statistics of the output of a different model of the same system. We use two biological examples to give a proof-of-principle demonstration of the methodology, and discuss its advantages  and potential further applications.

\end{abstract}

\section{Introduction} 

Computational modelling in the sciences is founded on the notion of abstraction, the process of identifying and representing mathematically the salient features and interactions of a real system. Abstraction is a human led and interdisciplinary activity: many factors influence the decision of which features/ interactions are eventually represented in the abstracted model, including the specialist interests of the scientists formulating the model, as well as computational constraints on the level of detail which can feasibly be implemented on the available hardware. Such factors inevitably vary between different research groups and at different times, leading to a proliferation of different models representing the same underlying phenomena.

Systems biology is a prominent field where models with different level of abstraction coexist. As an example, consider the process of gene expression, whereby genetic material stored in the DNA is transcribed into messenger RNA and eventually translated into protein. At the highest level of abstraction, which is frequently employed when studying high-throughput data, the process may be considered as a black box, and one may only be interested in characterising the statistical structures underlying observed levels of gene expression in different conditions \cite{hoyle2002making}. Alternatively, one may want to mechanistically represent the dynamics of some of the steps involved in the process. The choice of which steps to include is again largely driven by extrinsic factors: examples in the literature range from highly detailed models where the synthesis of mRNA/ protein is modelled through the binding and elongation of polymerase/ ribosomes, to models where protein production is modelled as a single reaction with first order kinetics \cite{aitken2013rule,lawrence2006modeling}.

Representing the same physical process by multiple models at different levels of abstraction immediately engenders the question of how different model outputs can be reconciled. As the models all represent the same underlying physical system, it can be plausibly assumed that such models will agree at least qualitatively for suitable parametrisations. In general, however, models may not agree quantitatively, and their discrepancy may be a function of the parameters. Understanding and quantifying such discrepancies would often be very valuable: first of all, it can shed light on how simplifications within models affect predictions, and secondly it may open the opportunity to construct computationally efficient surrogates of complex models. Such surrogates can be precious when modelling requires intrinsically computationally intensive tasks,  like  inference.

In this paper, we approach the problem of reconciling models from a statistical machine learning angle. We start by sampling a sparse subset of the parameter space over which we evaluate the models' outputs (generally by simulation). These evaluations are used as a {\it training set} to learn a {\it correction map} via a non-parametric regression approach based on Gaussian Processes. We show that our approach yields a consistent stochastic equivalence between models, which provably reconciles the predictions of the two models up to the second moment. We demonstrate the approach on two biological examples, showing that it can lead to non-trivial insights into the structure of the models, and provide an efficient way to simulate a complex model via a simpler model.

The rest of the paper is organised as follows: we start by giving a high level description of the problem and how we attack it. This is followed by a formal definition and a thorough illustration of the proposed solution, discussing its desirable theoretical properties. We then demonstrate the approach on two proof of principle examples, showing the potential for practical application of the method. We conclude the paper by discussing the relationship of our approach to existing ideas in model reduction and statistical emulation, as well as some possible extensions of our results.

\section{Problem definition}

\paragraph{High level description of the approach.}
In this paper we consider the problem of performing analyses which require exhaustive sampling of a model's outputs, $\bM$,  the dynamics of which are expensive to compute. 
We are not interested in the origin of such a complexity, and {\em we assume} to be given an {\em abstraction}/{\em surrogate}  of this model, $\bm$,  which is reasonably less challenging to analyze\footnote{$\bM$ could be complex to analyze  either because of its structure, e.g., it might have many variables, or its numerical hurdles, e.g., the degree of non-linearity or parameters stiffness. For   similar reasons, we do not care whether $\bm$ is has been derived by means of independent domain-knowledge or automatic techniques.}.  For this reason, {\em we want to investigate a general methodology to use $\bm$    as a reliable proxy to get  statistics over $\bM$.}  Possible   applications of this framework, which we discuss in  \S\ref{sec:applications}, regard {\em model selection and synthesis},  {\em inference}, and {\em process equivalence/control}.

In general, we will assume both models to be stochastic processes, e.g. continuous time Markov Chains (CTMCs). Furthermore, we assume that the highly detailed model $\bM$ and the reduced model $\bm$ share some parameters $\pmpar$ and some observables/ state variables $\cvar$, but the large model will generally have many more state variables and parameters.  In general we can compute some statistics of the shared state variables $\cvar$ (e.g. mean), and that such computation will be considerably more expensive using the detailed model $\bM$.

As both models are devised as abstractions of the same physical system, it is not unreasonable to assume that the expected values of the shared variables will be similar for some parametrisations of the models. However, it is in general not the case that the {\it distribution} of the shared variables implied by the two models will be the same, and, as we vary the shared parameters $\pmpar$, even the expected values may show non-negligible discrepancies. Our aim is to develop a generally applicable machine learning approach to correct the output of the reduced model, in order to match the distribution of the larger model. This has strong practical motivations: one of the primary uses of models in general is to test hypotheses statistically against observed data, and it is therefore essential to capture as accurately as possible the implied variability on observable variables.

The strategy we will follow is simple: we start by sampling values of the shared parameters $\pmpar$, and compute the first two statistics of the observed variables as implied by both the large and reduced models (by simulation). In general, one may expect the variance implied by the larger model to be larger, as a more detailed model will imply more stochastic steps. We can therefore correct the first two statistics (mean and variance) of the reduced model output by adding a random function of the shared parameters $ \bth_{\bm}$, which can be learned by rephrasing it as a regression task. We will work with heteroschedastic Gaussian Processes  \cite{lucaQEST13}.

\subsection{Formal problem statement}

We assume to be given two models of the same underlying physical system:
\begin{itemize}
	\item a highly detailed model $\bM$, with state variables $\mathcal{Y}$ and parameters $\pMpar$. 
	\item a reduced model $\bm$, with state variables $\mathcal{X}$ and parameters $\pmpar$. 
\end{itemize}
We will have $|\mathcal{Y}| \gg |\mathcal{X}|$ and $|\pMpar| \gg | \pmpar|$.

\paragraph{Assumptions.} In general, the problem we want to tackle draws immediate connection to that of using $\bm$  as a {\em statistical emulation} of $\bM$. However, to exploit solutions from {\em regression analysis} and {\em machine learning}, we make   further assumptions and discuss their limitations thorough the paper. 

\begin{enumerate}
\item ({\em Observables}) we assume that  it exists a non-empty  set of state variables (or, more generally, observables) ${\cvar}$, common to both models, that is sufficient to compute our statistics of interest.

\item ({\em Parameters}) we assume that model $\bM$ is fully  parametrized by  $\pMpar$, a vector of real-valued parameters that  breaks down as $\pMpar = [ \pmpar\quad \pfpar ]^\top$, with $\pmpar$ shared between models $\bm$ and $\bM$.
Here, we assume that $\bm$ is fully parametrized\footnote{In principle, even $\bm$ might have a set of free variables, with respect to $\bM$. However, as we have full control over that model, we could  assume a parametrization of such variables and all what follows would be equivalent.} by $\bth_{\bm}$, which ranges  in a domain set $\Theta$. We term $\pfpar$  free parameters in $\bM$, given $\bth_{\bm}$. We further assume to have a probability density $p(\pfpar)$ for the free parameters, and a probability density $p(\pmpar)$ for the shared parameters, encoding our knowledge about them.

\item ({\em Sampling}) we assume that, for every parametrization,   each model's dynamics is computable, i.e. it can be simulated. 
 \end{enumerate}

\begin{figure}[t]
\centerline{
\includegraphics[width=.8\textwidth]{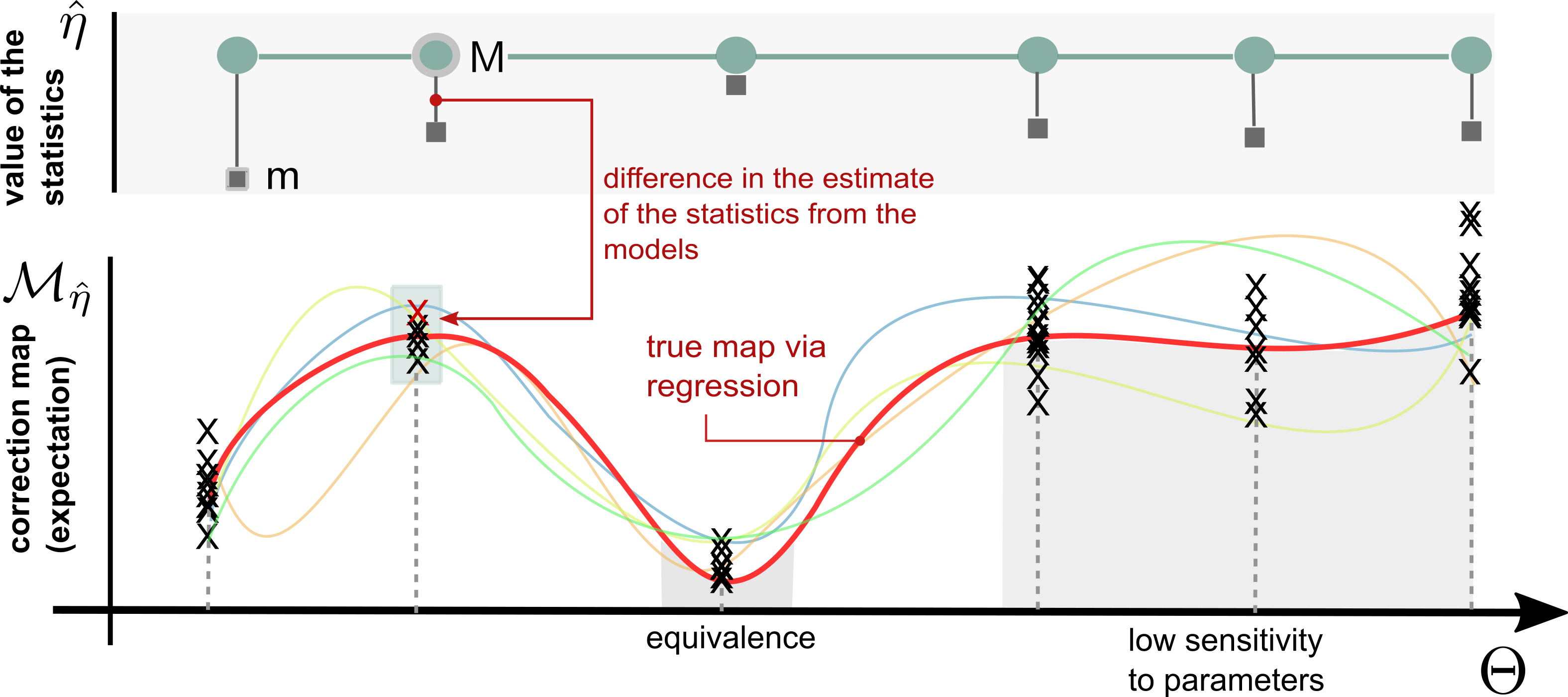}}
\caption{Correction maps as regression problems. From sampled estimates of a  statistics $\stat$ computed over  domain $\Theta$ by models $\bM$ and $\bm$, we can define a correction $\ebM_{\stat}(x)  \triangleq \bm_{\stat}(x) + {\map}(x)$ from their difference. According to the variables involved, we collect multiple values per parameter value; see also Figure \ref{fig:marg}. Then, a regression over such values  leads to a correction model  for $\bM$'s prediction, from $\bm$'s ones; this differs from standard emulation as we retain a mechanistic description of target system, via $\bm$. Maps highlight regions of the parameters where concordance among models and parameters' sensitivity varies, and can be retrieved with  regression schema that account for the heteroschedastic variance in the observations.}
\label{fig:model}
\end{figure}

In this work, we will correction maps conditioned to a reference statistic of interest, in the following sense.

\begin{mydef}[Statistic]
A statistic $\tstat$ is {\em any observable that we can compute from  one, or from an ensemble of simulations of a model}.  We denote its {\em estimator} computed from model $\sf{q}$ with parameters $x$ as  $\sf{q}_{\stat}(x)$, and its true value $\sf{q}_{\tstat}(x)$.
\end{mydef}
Valid examples of such statistics  are, e.g., a single value or the expectation of a  variable in $\cvar$, the satisfiability of a temporal logical formula depending on variables $\cvar$ that could be model-checked, etc. The  richer the estimator, the  higher number of samples are required for the estimator to converge to the true statistics. We will make use of estimators that are {\em consistent} in the following sense: $\sf{q}_{\stat}(x) \to \sf{q}_{\tstat}(x)$ as the number of samples goes to infinity.

\begin{mydef}[Correction map]
	\label{def:correction}
We define   {\em  an $\epsilon$-correction map} ${\tmap}: \Theta \to \R^w$ for a {\em statistics} $\tstat$ to be a function  that for any  $\bth_{\bm}\in \Theta$, satisfies 
{\small
\begin{align}\label{eq:correction}
&\ebM_{\tstat}(\bth_{\bm})  \triangleq \bm_{\tstat}(\bth_{\bm}) + {\tmap}(\bth_{\bm}) 
\;\text{and}\; \int_\Theta \parallel \ebM_{\tstat}(\bth_{\bm})  - \mathbb{E}_{\pfpar}[\bM_{\tstat}(\bth_{\bm})]\parallel_2 p(\pmpar)d\pmpar  <   \epsilon  
\end{align}}
where $\epsilon>0$ is the precision,  
and $\mathbb{E}_{\pfpar}[\bM_{\tstat}(\bth_{\bm})] = \int \bM_{\tstat}(\pmpar; \pfpar)p(\pfpar)d\pfpar$ is the {\em expectation of the statistics} computed from $\bM$, with respect to its free parameters $\pfpar$. $\ebM$ is our {\em corrected prediction} of $\bM$. 
\end{mydef}
 Thus, $\map$  can  correct the outcomes of $\tstat$ assessed over $\bm$, $\bm_{\tstat}(\bth_{\bm})$, to match (with a given precision) those that we would have by computing the statistics over $\bM$. Notice that the corrected outcome has no more dependence from the free parameters, as this is a correction in expectation with respect to $\pfpar$. 

\begin{SCfigure}[][t]
\includegraphics[width=.5\textwidth]{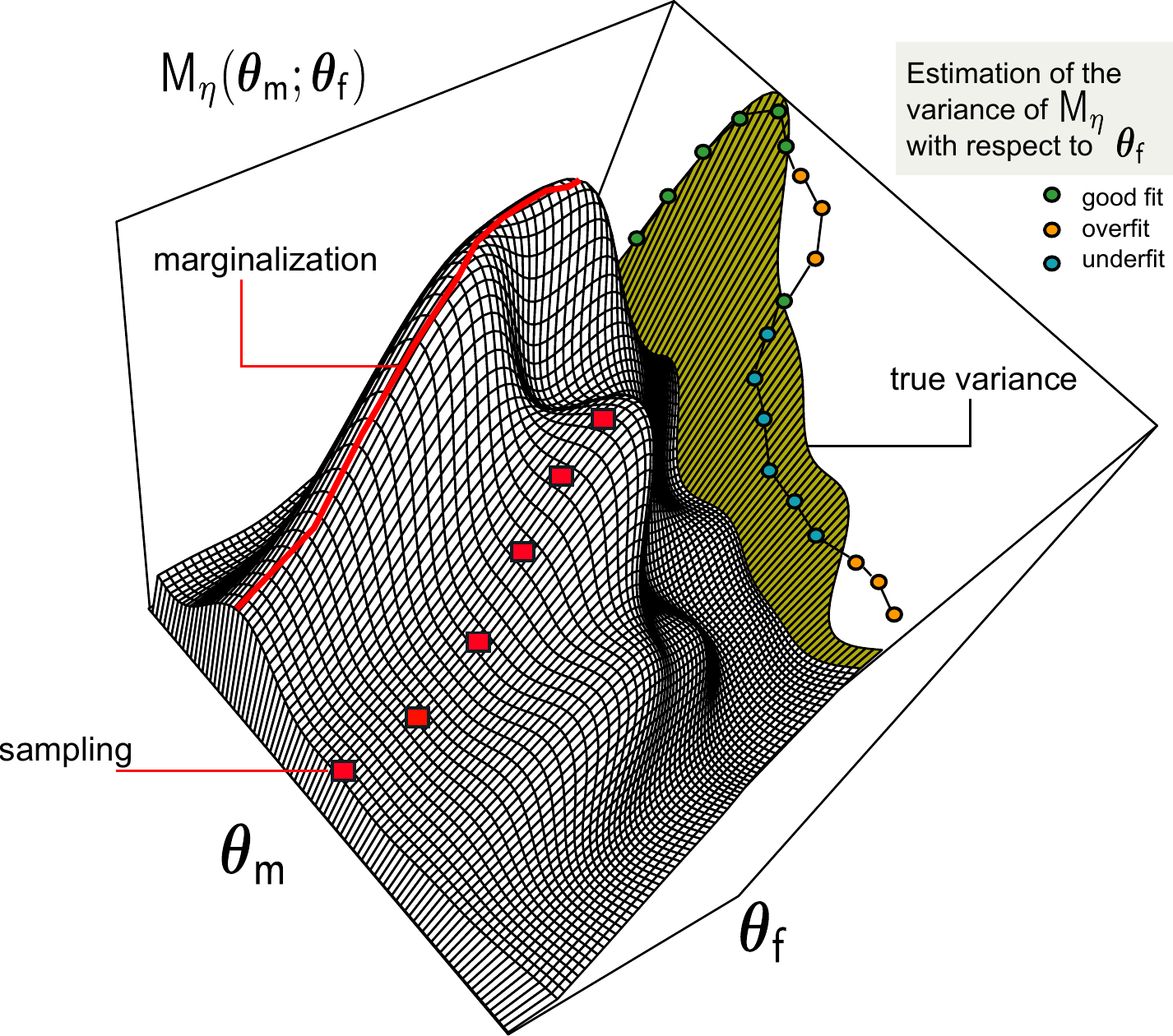}
\caption{The statistics computed from $\bM$, $\bM_{\tstat}(\bth_{\bm}; \pfpar)$, has a variance whose contribution depends on the parametrization of all model's variables, hence on both $\bth_{\bm}$ and $\pfpar$. Marginalization is the strategy that, at an exponential cost, ensures that our estimate of such variance is correct; all downsampling strategies, instead, could lead to over or under estimation of such variance. In any case, it is important to account for the variation in such variance as a function of $\bth_{\bm}$; this can be achieved by heteroschedastic regression.
}
\label{fig:marg}
\end{SCfigure}

Notice that the correction is a $w$-dimensional vector, and hence $\parallel \cdot \parallel_2$ is used as  distance metric, and that  the  term $\epsilon$ allows for tolerance in the correction's precision. It is easy to define the optimal, zero-error, correction map:
\begin{equation}\label{eq:marginal}
\tmap^\star(\bth_{\bm}) \triangleq 
\mathbb{E}_{\pfpar}[\bM_{\tstat}(\pmpar)]  - \bm_{\tstat}(\bth_{\bm}) \, .
\end{equation}
However, the correction function $\tmap^\star(\bth_{\bm})$ is impossible to compute exactly, as we cannot compute neither $\bM_{\tstat}$ nor its marginalisation over $\pfpar$. Hence, we will learn an approximation of $\tmap^\star(\bth_{\bm})$ trying to keep its error low so to satisfy Definition \ref{def:correction}.  We turn this problem into a {\em regression task}, as graphically explained in Figures \ref{fig:model} and  \ref{fig:marg}, and exploit Gaussian Processes.

\section{Learning the Correction Map}\label{sec:solution}
In this section we will present our machine learning strategy in more detail.  We consider the case of a scalar statistics, as $w$-dimensional ones can be treated by solving $w$ independent learning problems. 

\subsection{Marginalising $\pfpar$}
In order to evaluate \eqref{eq:marginal}, we need to be able to compute or approximate the value $\mathbb{E}_{\pfpar}[\bM_{\tstat}(\pmpar)]$ with respect to the free parameters of $\bM$, for a any given  $\pmpar$.  As this integral cannot be treated analytically or numerically, due to the large dimensionality involved (the cost is exponential in $|\pfpar|$), we will resort to statistical approximations. 
Before discussing them, let us comment on the distribution $p(\pfpar)$, which is an input for our method. In particular, this distribution encodes our knowledge on the more plausible values of the free parameters. In case we have no information, we can choose an uniform distribution. On the other side of the spectrum, we may know the true value $\pfpar^*$ of $\pfpar$, and choose a delta Dirac distribution, which will dramatically simplify the evaluation of the integral. In this case, we can evaluate \eqref{eq:marginal} as
\begin{equation}\label{eq:constant}
\tmap(\bth_{\bm}) \triangleq 
\bM_{\tstat}(\bth_{\bm}; \pfpar^*) - \bm_{\tstat}(\bth_{\bm}) \, ,
\end{equation}

Moreover, the approximation,
$\int \bM_{\tstat}(\bth_{\bm}; \pfpar)p(\pfpar) d\pfpar \approx \bM_{\tstat}(\bth_{\bm};\pfpar^*)
$ 
is appropriate when the distribution $p(\pfpar)$ is tightly concentrated around its mode $\pfpar^*$. 

In general, however,  when $p(\pfpar)$ does not have this special form, we can resort to  downsampling $\mathbb{E}_{\pfpar}[\bM_{\tstat}(\pmpar)]$, by generating $k$  samples ${\pfpar}^{(1)}$, $\ldots$, ${\pfpar}^{(k)}$ and approximating   $\mathbb{E}_{\pfpar}[\bM_{\tstat}(\pmpar)] \approx \frac{1}{k} \sum_j \bM_{\tstat}(\pmpar;{\pfpar}^{(j)})$. In the following, however, we will not necessarily aggregate the values $\bM_{\tstat}(\pmpar;{\pfpar}^{(j)})$, and treat them individually to better account for the variance in the observed predictions.

\subsection{Gaussian Processes} We will solve the learning problem resorting to Gaussian Process (GP) regression \cite{RASMUSSEN}. GPs are random functions, i.e. probability distributions over a function space, in our case functions $f:\Theta\rightarrow\mathbb{R}$, with the property that any finite dimensional projection $f(\bth_1)$, $\ldots$, $f(\bth_k)$ is a multidimensional Gaussian random variable. It follows that GP are defined by a mean function $\mu(\bth)$, returning the mean at any point in $\Theta$, and by a covariance or kernel function $k(\bth_1,\bth_2)$, for giving the covariance between any pair of points in $\Theta$.  GP can be used to solve regression tasks in a Bayesian setting. The idea is as follows: we put a GP prior on  the space of functions $\{f~|~f:\Theta\rightarrow\mathbb{R}\}$, typically by assuming constant zero mean and fixing a kernel function\footnote{ In this work, we use the classic Gaussian kernel fixing  hyperparameters by maximising the type-II likelihood; see \cite{RASMUSSEN}.}, and then consider the posterior distribution given a set of observations $Y= \{ y^{(i)}\}_I$, $i \in I$, at input points $X = \{\bth_{\bm}^{(i)}\}$.   If we assume that $y^{(i)} = f(\bth_{\bm}^{(i)}) + \epsilon_i$, with $\epsilon_i$ a zero mean Gaussian noise term with variance $\sigma^2$, then we  obtain that the posterior distribution given the observed data is still a GP, with mean and kernel functions that can be obtained analytically, see \cite{RASMUSSEN} for further details. GP regression is essentially the same if the observation noise $\sigma^2$ is different at different input points, i.e. $\sigma^2 = \sigma(\rarg{i} )^2$, in which case we talk about heteroschedastic regression. 

\subsection{The regression task}

Let  ${\bth_{\bm}^{(i)}}$ for some index set $i \in I$ be the {\em input space}, and  $\{ \langle {\bth_{\bm}^{(i)}}, y^{(i)}\rangle\}_I$  our {\em training points}.   In case we use eq.~\eqref{eq:constant} to evaluate the correction map,  each $y^{(i)}$ is a scalar  value, and a standard regression schema based on Gaussian Processes can be used. In that case we assume samples of the correction map $y$ to be observations from a random variable centered at a value given by the {\em latent function}
\begin{equation}\label{eq:reg}
y^{(i)} \sim {\cal N} (\tmap(\rarg{i}),    \sigma^2) \, .
\end{equation}
In this standard  Gaussian Processes regression the noise model in the observations is assumed to be a constant $\sigma^2$ for all sampled points.

In the more general case we work with downsampling solutions that exploit  $k$  samples for the free variable, ${\pfpar}^{(1)},\ldots, {\pfpar}^{(k)}$. In that case, we have  $k$  correction values per training point, 
$
\Big\{ \langle {\bth_{\bm}^{(i)}}, [y^{(i,1)}  \cdots y^{(i,k)}]^\top \rangle\Big\}_I
$,
that we can use in a straightforward way to reduce to the above schema, or to estimate the variance of $\bM$ conditioned to its free variables. In these cases, the training set is $\{ \langle {\bth_{\bm}^{(i)}}, \overline{y}^{(i)} \rangle \}_I$,
namely we do regression on the point-wise expectation of the correction (i.e., $\overline{y}^{(i)}= \frac{1}{k}\sum_{j=1}^k y^{(i,j)}$).

\begin{est}[Empirical $\overline{\sigma}$-estimator.] Set $\overline{\sigma}$ to the empirical estimate of the variance {\em across all correction values}  to exploit  the schema in eq. (\ref{eq:reg}) with $\sigma^2 \triangleq \overline{\sigma}$.
\end{est}

Besides  the simple first case, it  is more interesting to account for a {\em model of the variance in the observations of the predictions from a model}; we discuss how this could be done in two ways.

\begin{est}[Point-wise $\sigma$-estimator] Let $\sigma(\cdot)$ the {\em empirical estimator} of the variance of the correction values, per training-point
\begin{equation}\label{eq:empvar}
\sigma(\rarg{i}) \triangleq \var[y^{(i,1)}, \ldots, y^{(k,1)}] \, ,
\end{equation}
then define a model of the variance as a point-wise function of the regression parameter, that is
\begin{equation}\label{eq:sreg}
y^{(i)} \sim {\cal N}\Big (\tmap(\rarg{i}),   \sigma(\rarg{i})^2\Big) \, .
\end{equation}
\end{est}
In this case, the variance that we expect in each prediction of the latent function is  estimated from the data, thus leading to a form of {\em heteroscedastic} regression.

We can estimate with higher precision a model of the variation in the variance across the input space; to do that we  perform {\em regression of the variance change}, and then inform the outer regression task of that prediction.

\begin{est}[Nested $\sigma$-estimator] Consider the same estimator of the variance as above,  and collect the variance estimates as $\{ \langle {\bth_{\bm}^{(i)}}, w^{(i)} \rangle\}_I$.  Learn  a latent function model of the true variance $\sigma_\star(\cdot)$, which we assume to have a fixed variance $\sigma_\star^2$
\begin{align}\label{eq:regvar}
w^{(i)} \sim {\cal N}\Big ( \sigma_\star(\rarg{i}),   \sigma_\star^2 \Big) &&
y^{(i)} \sim {\cal N}\Big (\tmap(\rarg{i}),   \sigma_\star(\rarg{i})^2\Big) \, .
\end{align}
\end{est}
This is also a form of heteroschedastic regression, but nesting two GP regressions to account in a finer way for the variance's changes.

\subsection{Properties of the correction map}

The correction map that we learn, for all variance schemes, is a statistically sound estimator of $\mathbb{E}_{\pfpar}[\bM_{\tstat}(\pmpar)]$, in the sense of being consistent. 

\begin{myth}[Correctness] Let $\bm_{\stat}(\bth_{\bm})$ and $\mathbb{E}_{\pfpar}[\bM_{\stat}(\pmpar)]$ be consistent estimators of $\bm_{\tstat}(\bth_{\bm})$ and $\mathbb{E}_{\pfpar}[\bM_{\tstat}(\pmpar)]$, then $\ebM_{\stat}(\bth_{\bm})  \triangleq \bm_{\stat}(\bth_{\bm}) + {\map}(\bth_{\bm})$ is a consistent estimator of  $\mathbb{E}_{\pfpar}[\bM_{\tstat}(\pmpar)]$, for any estimation strategy of $\map$.
\end{myth}
The result follows because $\map$ converges to $\tmap$ due to properties of GPs \cite{RASMUSSEN}, and because of the consistency of $\bm_{\stat}$ and $\mathbb{E}_{\pfpar}[\bM_{\stat}(\pmpar)]$. The proof is sketched in Appendix \ref{app:proofs}.

The correction map ${\map}(\bth_{\bm})$ is estimated from samples of the system, hence it is  an approximation of the exact map defined by eq. \eqref{eq:marginal}. Thus, it is a correction map in the sense of Definition \ref{def:correction}. However, being the result of a GP regression, ${\map}(\bth_{\bm})$ is in fact a random function. Therefore, in measuring the error according to eq. \eqref{eq:correction}, we need to consider the average value of the random function $\mathbb{E}[{\map}(\bth_{\bm})]$. The variance  $\var[{\map}(\bth_{\bm})]$, instead, provides a way of computing the error $\epsilon$, but in a statistical sense with confidence $\alpha$: $\epsilon$ can be estimated by averaging over $\Theta$ (with respect to $p(\pmpar)$) the half-width of the region containing $\alpha$\% of the probability mass of  ${\map}(\bth_{\bm})$.

The cost of all  our approaches  inherently depends on how many samples we pick from $\Theta$, the way parameters in $\pfpar$ are accounted for and the number of parameters in $\pmpar$. The sampling cost in general grows exponentially with $|\pmpar|$, and each Gaussian regression is cubic in the number of sampled values. Notice that, asymptotically, the cost of the empirical and nested $\sigma$-estimators is the same,  as the two regressions are executed in series.

\section{Applications}
\label{sec:applications}

We discuss now several potential applications of our framework. The advantages of using our approach are mostly computational: the reduced model is simpler to analyze, yet it retains a mechanistic interpretation, compared to statistical emulation. 
\paragraph{Model Building.} 
Many common problems in the area of dynamical systems can be tackled by resorting to correction maps. 
\begin{myprob}[Model selection]  Let $\bM$ be a model, and $\bm_1$, $\ldots$, $\bm_k$ a set of candidate models for correction, each one having a correction map $\tmapi{1}$, $\ldots$, $\tmapi{k}$.  The smallest-correction model $\bm^\ast$  for a statistic $\tstat$  is
$ \bm^\ast \triangleq \arg\min_{\bm_i} \int \tmapi{i}(\boldsymbol{\theta}) d\boldsymbol{\theta}$.
\end{myprob}
This criterion is certainly alternative to structural Bayesian approaches \cite{barber2012}, which can be used to select the structurally smaller model {\em within} an equivalence class (see below). Also, allows to frame a model synthesis problem.

\begin{myprob}[Model synthesis] For a model $\bM$ with parameters $\pMpar$ and for a statistic $\tstat$: $(i)$  partition $\pMpar$ into sets $\pmpar$ and $\pfpar$,  $(ii)$ generate a finite set of plausible reduced models with parameters $\pmpar$ and  $(iii)$ select the best one, according to the above model selection criterion.
\end{myprob}

In this case, the partition might aim at identifying in $\pMpar$  the model's parameters which contribute the most  to the  variance for the statistics $\tstat$. 
Opportunities for {\em control} are also plausible if one can reduce the problem of controlling $\bM$ to ``controlling and correcting'' a reduced model $\bm$. This should be easier as $\bm$ is structurally smaller than $\bM$, and so is lower the complexity of solving a {\em controller synthesis problem}.

\paragraph{Parameter Estimation.} 
Another application of our framework is in Bayesian parameter estimation of the parameters $\pmpar$ of the big model $\bM$, given observations $D$ of variables $\cvar$, using the small model and the corrected statistics to build approximations of the likelihood function $p(D\mid \pmpar)$ to plug into a Bayesian approach (i.e. to compute an approximate posterior). For  instance, using the mean of variables $\cvar $ (and the variance of the correction map as a proxy of the variance $\cvar$ in $\bM$ after marginalisation of $\pfpar$), we can easily compute the distribution of $\cvar$ under a linear noise approximation and use a Bayesian inference scheme such as\cite{komorowski2009}.

\paragraph{Model Equivalence.} Correction maps can also be used to compare processes, via weak forms of  {\em bisimilarity} with respect to the observations and a statistics. \begin{mydef}[Model equivalence] Models $\bM_1$ and $\bM_2$, for a statistic $\tstat$ and parameter sets $\pmpar$ and $\pfpar$, are {\em $\tstat$-bisimilar conditioned to $\bm$},  $\bM_1\equiv_{\bm}^{\tstat} \bM_2$, if and only if for all $\pmpar\in \Theta$, it holds $\tmapi{1}(\pmpar) = \tmapi{2}(\pmpar)$. A class of equivalence of models  with respect to $\bm$ and $\tstat$ is the set of all such bisimilar models.
\end{mydef}
This notion of bisimilarity resembles conditional dependence, as  we are saying that two models are equivalent if  an observer    corrects both the same way.  In this case, $\bm$ is a {\em universal corrector} of $\equiv_{\bm}^{\tstat}$, as it can correct for all the models in the class. The class of models that are equivalent to a model $\bM$ is $\{ \bM^\ast \mid  \bM^\ast \equiv_{\bM}^{\tstat}  \bM\}$ -- i.e.,  the class of models with  correction zero; notice that in this case $\pfpar=\emptyset$.  The previous definition can be relaxed by asking that $|\tmapi{1}(\pmpar) - \tmapi{2}(\pmpar)|\leq \epsilon$,  for all $\pmpar\in \Theta$.

\begin{remark} Criterion $\equiv_{\bm}^{\tstat}$ is a weaker form of {\em probabilistic bisimilarity},  namely if $\bM_1 \equiv \bM_2$ are  bisimilar, then   $\bM_1 \equiv_{\bm}^{\tstat}  \bM_2$ for some $\bm$ and any statistics of interest.  For instance, for CTMCs, this follows because $\equiv$ implies that $\bM_1$ and $\bM_2$ have the same infinitesimal generators for any parameter $\pmpar$ and $\pfpar$, hence the outcomes of $\bM_1$ and $\bM_2$ are indistinguishable, and  so are their statistics. 
\end{remark}

\begin{figure}[t]
\begin{center}
\begin{tabular}{>{\centering\arraybackslash}p{6.0cm}>{\centering\arraybackslash}p{4.0cm}}
{\em Full model} ($\bM$) & {\em Reduced model} ($\bm$) \\\hline
\end{tabular}\\
\begin{tikzcd}
\E + \SU\arrow[bend left=15]{r}{k_1} & \ES \arrow[bend left=15]{l}{k_{-1}} \arrow{r}{k_2} & \E{} + \PR{}
\end{tikzcd} $\quad\quad\quad\quad$ 
  \begin{tikzcd}
	\SU\arrow{r}{f} &  \PR{}
\end{tikzcd} \\
\begin{tabular}{>{\centering\arraybackslash}p{6.0cm}>{\centering\arraybackslash}p{4.0cm}}
 &  \\\hline
\end{tabular}\\
\begin{tikzcd}
 \GEA^\ast \arrow{r}{\alpha}\arrow[bend left=25]{d}{k_{\text{\sf off}}\PR{}} & \RNA^\ast\arrow{r}{\beta}\arrow{d}{\delta_{\text{\sf RNA}}} & \PR \arrow{d}{\delta_{\text{\sf P}}}  \\
  \GEI \arrow[bend left=25]{u}{k_{\text{\sf on}}} & \varnothing & \varnothing
\end{tikzcd} $\quad\quad\quad\quad$ 
\begin{tikzcd}
 \GEA^\ast \arrow{r}{\beta}\arrow[bend left=25]{d}{k_{\text{\sf off}}\PR{}} & \PR \arrow{d}{\delta_{\text{\sf P}}}  \\
  \GEI \arrow[bend left=25]{u}{k_{\text{\sf on}}} & \varnothing 
\end{tikzcd} 
\end{center}
\caption{Example models tested in this paper. Top panel: the Henri-Michaelis-Menten model, where $\bm$ is derived when  $\COND:  [\E{}]_0 + [\ES{}]_0 \ll [\SU{}]_0 + K_{\sf MM}$. Bottom panel, a protein translation network where $\bm$
when $\COND: \beta \gg \alpha$.}
\label{fig:ex-models}
\end{figure}
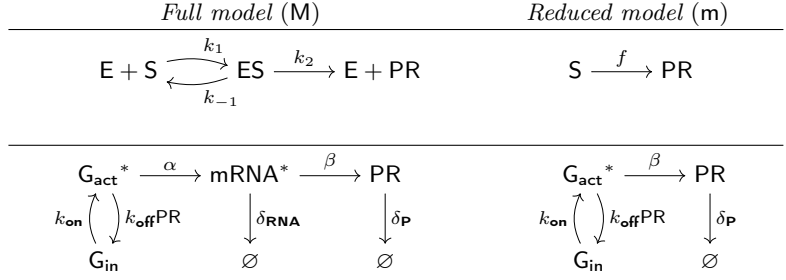

\section{Examples}

We investigate two examples to better illustrate our method.

\subsection{Model reduction via \QSSA{}} \label{sec:examp1}

Consider the irreversible canonical enzyme reaction with its exact representation (here,  model $\bM$),  for enzyme \E{}, complex \ES{}, substrate \SU{} and product \PR{} (Figure \ref{fig:ex-models}, top left panel).
When the concentration of the intermediate complex does not change on the time-scale of product formation, product is  produced at rate $f \triangleq {V_{\sf M}\SU{}}/(K_{\sf M} + \SU{})$ where $V_{\sf M} = k_2([\E{}]_0 + [\ES{}]_0) $ and $K_{\sf MM}=({k_{-1} + k_2})/{k_1}$.
This  is     the    Henri-Michaelis-Menten kinetics 
and is technically derived by a quasi-steady-state assumption (\QSSA{}), i.e., $\deriv{\ES{}}= \deriv{\E{}}=0$, that is in place when
$\COND:  [\E{}]_0 + [\ES{}]_0 \ll [\SU{}]_0 + K_{\sf MM}$, where $[x]_0$ is the initial amount of species $x$.  $\bm$ is thus the \QSSA{} reduced model (Figure \ref{fig:ex-models}, top right panel).

\begin{figure}[t]\center
\includegraphics[width=1\textwidth]{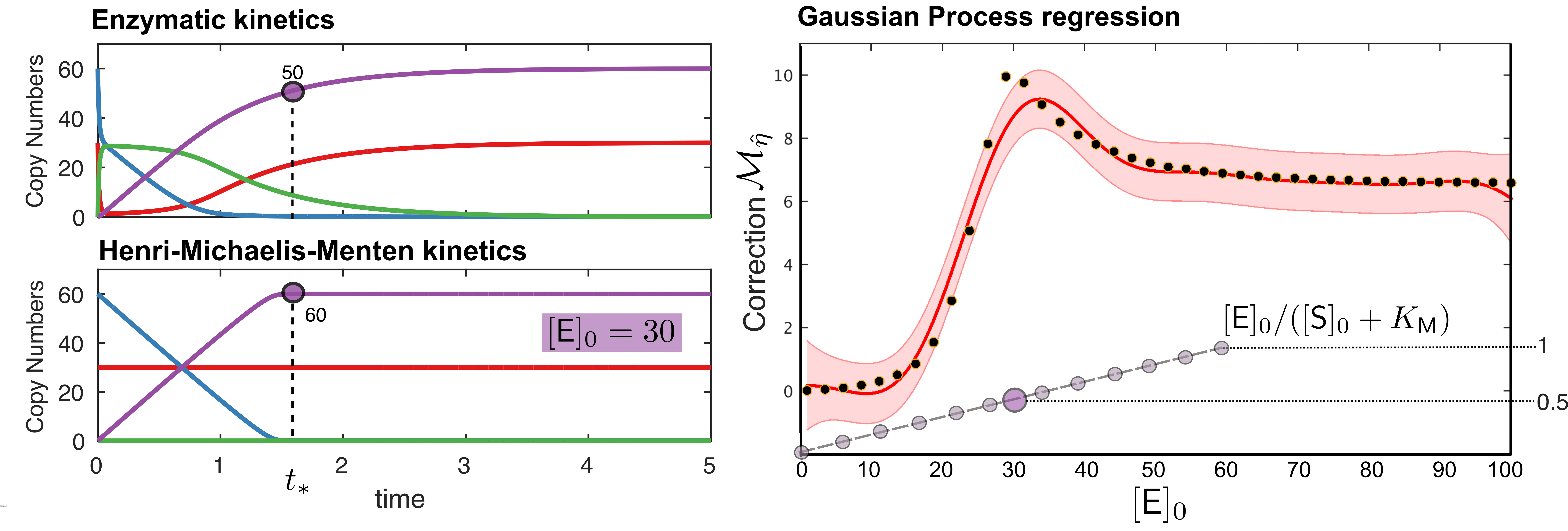}
\caption{Correction of the product formation at the transient time $t_\ast=1.5$, for a mean field model of irreversible canonical enzyme reaction and its simplified Henri-Michaelis-Menten form. Here  $k_1  = 2$, $k_{-1}=1$ and $k_2  = 1.5$, $[S]_0=60$ and $[P]_0=0$.  Regression over $[\E{}]_0$ is done with $40$ training points from $(0,100]$, and the correction in eq. (\ref{eq:constant})  as $\bM$'s free variables are part of the Michaelis-Menten constant.}
\label{fig:ex-mm}
\end{figure}

We interpret these two models as two systems of {\em ordinary differential equations}, see Appendix \ref{sec:app_ex}, and learn a correction for the following statistics
\begin{align}
\tstat &: \mathbb{E}[\PR(t_\ast)]  \text{, with }  \PR(t_\ast) \text{ the number of products at time } t_\ast
\end{align}
For non-degenerate parameter values both models predict the same equilibrium state, where a total transformation of substrate into product has happened, $\mathbb{E}[\PR(t)]   \to [\SU{}]_0$ for large $t$.  Thus,  we are not interested in correcting the dynamics of $\bm$ for long-run times, but rather in the transient (i.e., for small $t_\ast$). 

Also,  as  the \QSSA{} hypothesis  does not hold for certain initial conditions, we set $\pmpar = \{ [\E{}]_0\}$ as the regression variable, and set $[S]_0=60$ and $[P]_0=0$. The other parameters are  $k_1  = 2$, $k_{-1}=1$ and $k_2  = 1.5$  with unit {\em (mole per second)}$^{-1}$. In terms of regression, we pick $40$ samples of the initial enzyme amount from $(0,100]$, and set $t_\ast = 1.5$ as it is a time in the transient (manual tuning). This particular example is rather simple, as the free parameters of $\bM$  are part of the Michaelis-Menten constant and fixed, so we use the simpler correction of eq. (\ref{eq:constant}). Also, knowing when the \QSSA{}  holds  gives us an  interval where we expect the correction to shrink towards zero. The map is shown in Figure \ref{fig:ex-mm}, which depicts the expected concordance among the correction map and validity of the \QSSA{}.

\subsection{Model reduction via time-scale separation} \label{sec:examp2}

\begin{figure}[t] \center
\includegraphics[width=.9\textwidth]{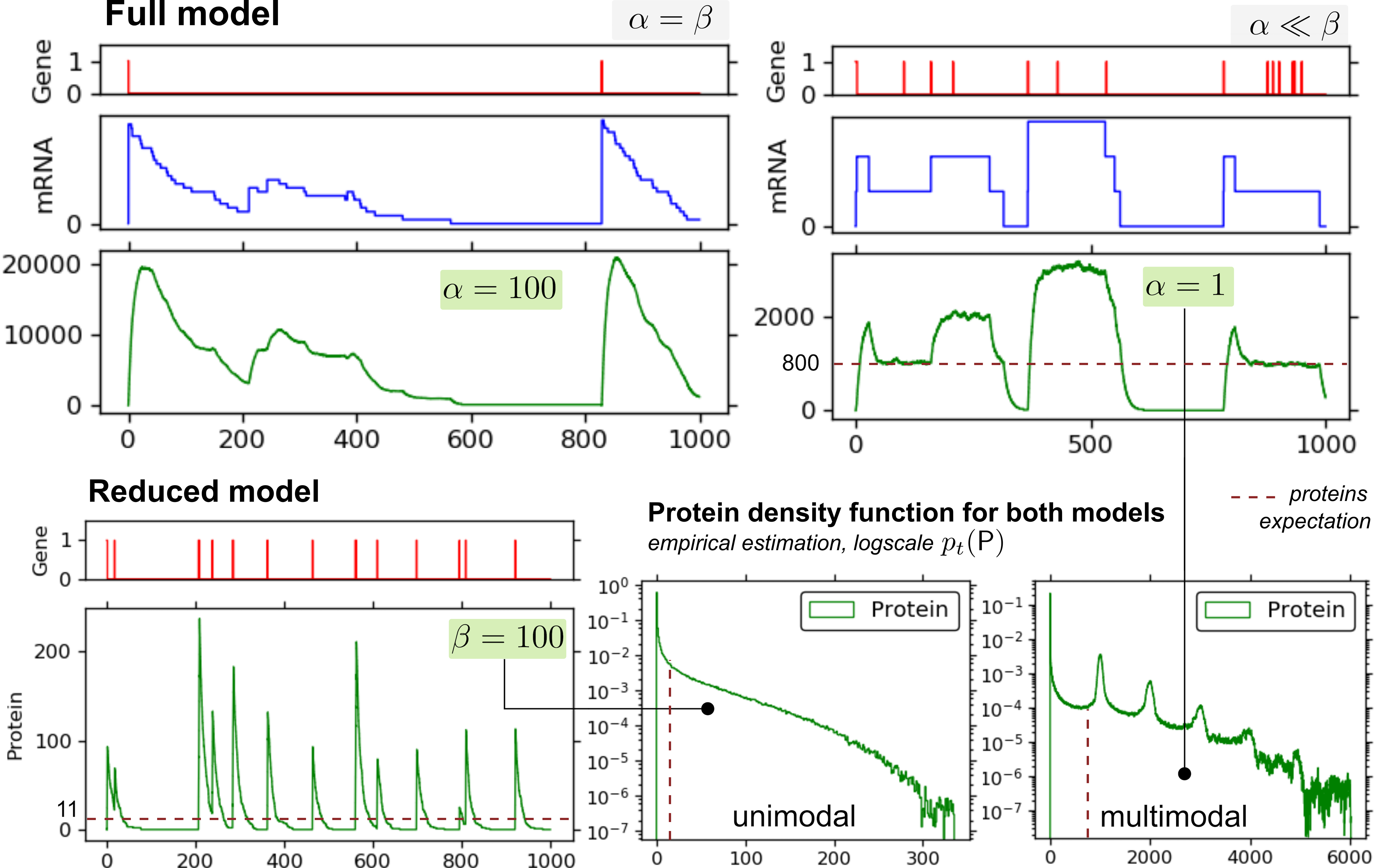}
\caption{Comparison between the dynamics of the full and the reduced models from \S \ref{sec:examp2}. Parameters are $k_{\text{\sf off}} = k_{\text{\sf on}} = 10^{-2}$, $\delta_{\text{\sf RNA}} =10 \delta_{\text{\sf P}} = 10^{-2}$, $[\GEA]_0=1$. Values for transcription ($\alpha$) and translation ($\beta$) are reported in the figure. The analysis of empirical proteins' distributions highlights the emergence of multimodal equilibria; the expectation on the number of proteins in the long run spans over different order of magnitudes, according to the relation between $\alpha$ and $\beta$.}
\label{fig:ex_PTN}
\end{figure}

Consider a  gene switching among active and inactive states of \RNA{} transcription  to be ruled by a {\em telegraph process} with rates $k_{\text{\sf off}}$/$k_{\text{\sf on}}$. A reaction model of such gene $\GE{}$, protein $\PR{}$, messenger \RNA{} with  transcription/translation  rates $\alpha$/$\beta$ as in Figure \ref{fig:ex-models}, bottom left panel.

Here the gene switches among {\sf active} and {\sf inactive} states, with rates $k_{\text{\sf on}}$ and $k_{\text{\sf off}}$, and \PR{} feedbacks positively on inactivation. Proteins and \RNA{}s are degraded with rates ${\delta_{\text{\sf P}}}$ and ${\delta_{\text{\sf RNA}}}$.
 In the uppermost part of the diagram species marked with a $\ast$ symbol are not consumed by a reaction, i.e., \RNA{} transcription is $\GEA \to \GEA +\RNA{}$. This model can be easily encoded in a Markov process, as discussed in \ref{sec:app_ex}.

We can derive an approximation to $\bM$  where the transcription step is omitted. This is valid  when $\COND:\beta \gg \alpha$ ({\em time-scales separation}), namely for every copy of \RNA{} a protein is immediately translated.

\paragraph{Correction of protein dynamics.} We build a correction map  with $\pmpar=\{\beta\}$. In this case the telegraph process is common to both models, but $\alpha$  and $\delta_{\text{\sf RNA}}$ are free variables of $\bM$; here we assume to have a prior delta distribution over the values of \RNA{}'s degradation,  so we set $\pfpar = \{\alpha\}$. For some  values of the transcription rate condition \COND{}  does not hold; in this case it is appropriate to account for $\alpha$'s contribution to the variance in the statistics that we want to correct, when we do a regression over $\beta$. Note that also  $\beta$ is part  of $\COND{}$.

Model  $\bM$ leads to stochastic bursts in \PR{}'s expression  when the baseline gene switching is slower than the characteristic times of translation/transcription. Here we set $k_{\text{\sf off}} = k_{\text{\sf on}} = 10^{-2}$, and assume \RNA{}'s lifespan  to be longer than protein's ones (also in light of condition \COND{}), so  $\delta_{\text{\sf RNA}} =10 \delta_{\text{\sf P}} = 10^{-2}$. We simulate both models with one active gene, $[\GEA]_0=1$; example dynamics are shown in  Figure \ref{fig:ex_PTN}, for $\beta=100$ and $\alpha\in\{1,100\}$. We observe that, when $\COND{}$ does not hold ($\alpha=\beta$) the protein bursts increases of one order of magnitude, and  the long-run  probability density function for the proteins, $p_{t}(\PR)$, becomes multimodal.

\newcommand{\eventually}[2]{\mathbf{F}_{[#1,#2]}}

We define  two statistics. One measures  the {\em first  moment} of the random variable that models the number of proteins in the long run; the other is a metric interval temporal logic formula \cite{alur1996}, expressing the probability of a protein burst  within the first 100 time units of simulation. 
\begin{align} \label{eq:stat-PTN}
\tstat_1 &: \mathbb{E}[\PR(t_\ast)]  \text{, with }  \PR(t_\ast) \text{ the number of proteins at time } t_\ast \gg 0 \\
\tstat_2 &: \mathbb{E}[p(\varphi)]  \text{, with }  \varphi \triangleq \eventually{0}{100} \PR(t) > 200
\end{align}
The former is evaluated by a unique long-run simulation of the model, as its dynamics are ergodic. For the latter we estimate the {\em satisfaction probability of the formula} via multiple ensembles, as in a parametric {\em statistical model checking problem}~\cite{smoothed}.

\begin{figure}[t]
\includegraphics[width=1\textwidth]{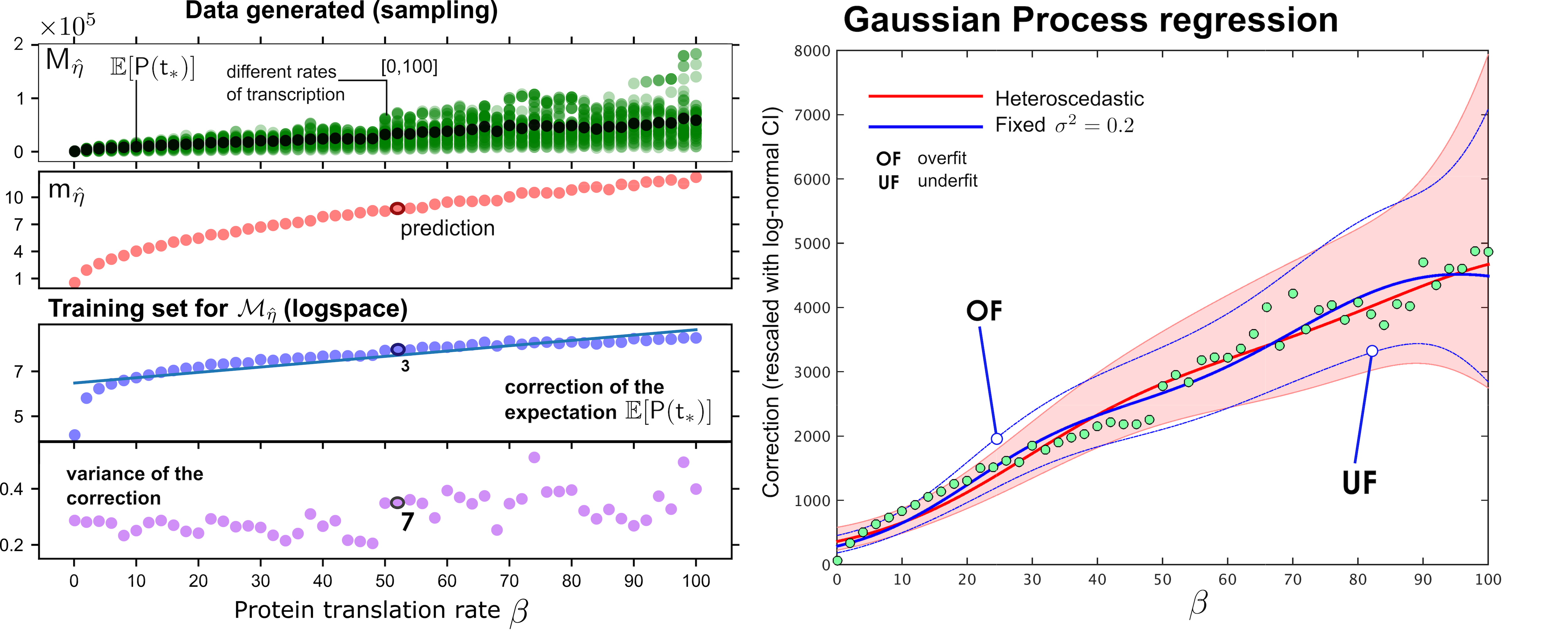}
\caption{For the regression task we sample $50$ values of $\beta$ from $[0.1, 100]$. For $\alpha$, instead, we sample $50$ random values in the same interval, for each value of $\beta$. Correction values are transformed logarithmically (left panel) before doing regression; observe that, on the linear scale, the correction is of the order of $10^3$ with variance $10^7$  (midpoint value $\beta \approx 50$). Gaussian Process regression (right panel) is performed with a constant $\sigma^2=0.2$, eq. (\ref{eq:reg}) and with the  $\sigma$-estimator, eq. (\ref{eq:sreg}). Values are re-scaled linearly, and 95\% log-normal confidence intervals are shown; regression highlights that  the posterior variances are similar, but  the fixed-variance schema tends to underfit or overfit the heteroscedastic variance ({\em assumed it to be closer to the true one}).
}
\label{fig:TR-prot}
\end{figure}

For the regression task we sample $50$ values of $\beta$, in the range $[0.1, 100]$. For $\alpha$, instead, we sample 50 random values in the same interval, for each value of $\beta$; notice that with high probability we pick values where $\COND{}$ does not hold, so we might expect high correction factors.  Data generated and the regression results are shown in Figure \ref{fig:TR-prot}, for both fixed-variance regression, empirical $\sigma$-estimator in eq. (\ref{eq:reg}) and with the  $\sigma$-estimator, eq. (\ref{eq:sreg}). Because variance spans over many orders of magnitude, regression is performed in the logarithmic space. Results highlight a general difference between the posterior variance between the two estimators.

 \begin{SCfigure}[][t]
\includegraphics[width=.65\textwidth]{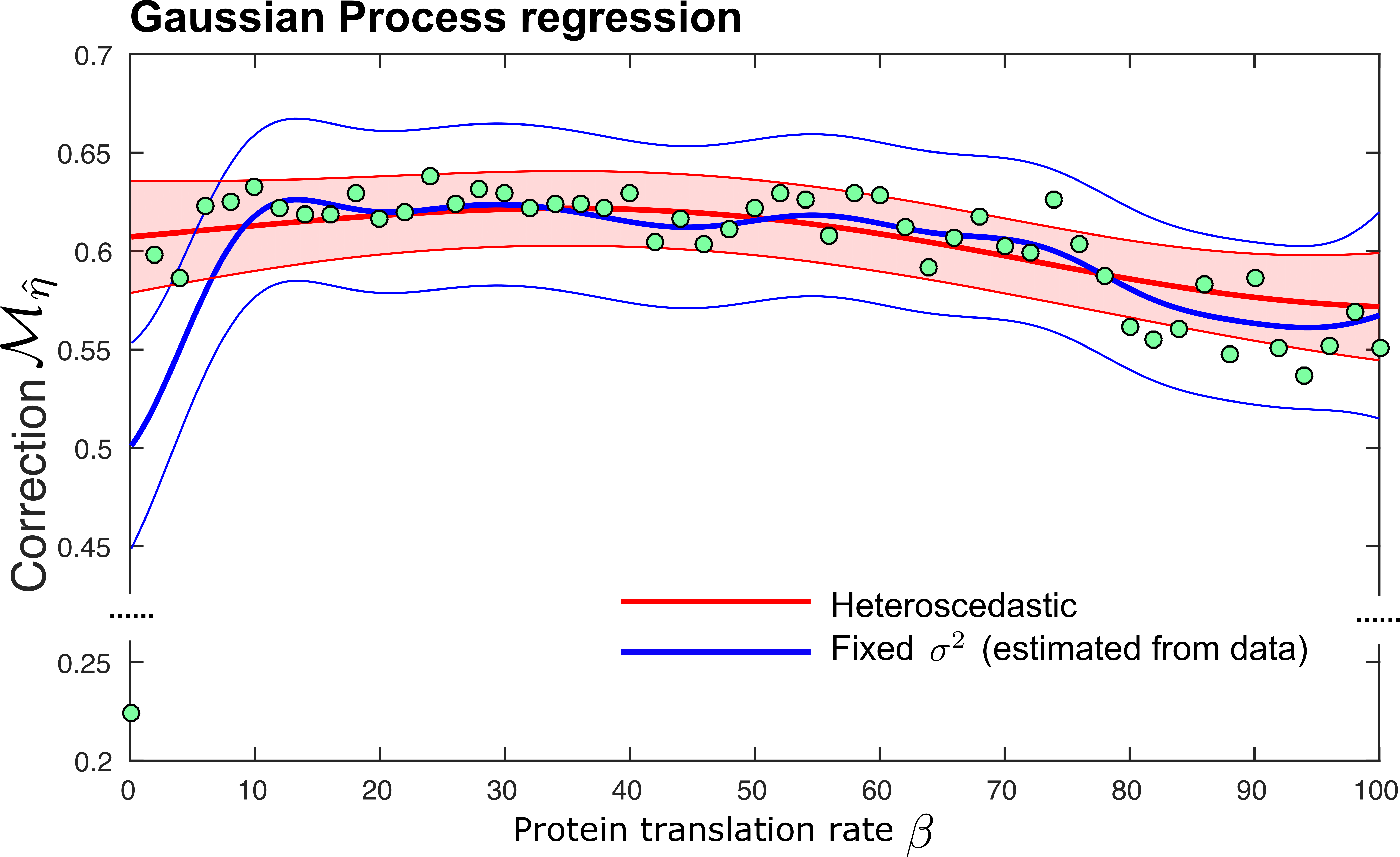}
\caption{Correction map for the expected satisfaction probability of the linear logic formula $\tstat_2$ in eq. (\ref{eq:stat-PTN}). Comparison between the point-wise $\sigma$-estimator and the empirical $\overline{\sigma}$-estimator. Sampled data is in Appendix Figure \ref{fig:TR-prob-data}.}
\label{fig:PTN_SAT_PROBAB_reg}
\end{SCfigure}
 
For the second statistics, data is generated from an initial condition where one gene is inactive, $[\GEI]_0=1$. Notice that the expected time for the gene to trigger its activation is $1/k_{\text{\sf on}} = 100$ (the time upper-bound of the formula), so for some parametrization there will be non-negligible probability of having no protein spike above threshold $200$. The formula satisfaction probability is evaluated with 100 independent model runs, and data generated are shown in Appendix Figure \ref{fig:TR-prob-data}. Regression results are shown in Figure \ref{fig:PTN_SAT_PROBAB_reg}, where the point-wise $\sigma$-estimator and the empirical $\overline{\sigma}$-estimator are compared, highlighting the more robustness of the former with respect to the sampled bottom-left outlier.

\section{Conclusions}

Abstraction represents a fundamental tool in the armoury of the computational modeller. It is ubiquitously used in biology as an effective mean to reduce complexity, however systematic analysis tools to quantify discrepancies introduced by abstraction are lacking. Prominent examples, beyond the ones already mentioned, include models with delays, generally introduce to approximately lump multiple biochemical steps \cite{caravagna2011formal}, or physiological models of organ function which adopt multi-scale abstractions to model phenomena at different organisational levels. These include some of the most famous success stories of systems biology, such as the heart model of \cite{noble2002modeling}, which also constitutes perhaps the most notable example of a physical systems which has been modelled multiple times at different levels of abstraction. Employing our techniques to clarify the quantitative relationship between models of cardiac electrophysiology would be a natural and very interesting next step.

Our approach has deep roots in the statistical literature. In this paper we have focussed on the scenario where we try to reconcile the predictions of two separate models, however the complex model was simply used as a sample generator black box. As such, nothing would change if instead of a complex model we used a real system which can be interrogated as we vary some control parameters. Our approach would then reduce to fitting the simple model with a structured, parameter dependent error term. This is closely related with the use of Gaussian Processes in the geostatistics literature \cite{cressie2015statistics}, where simple (generally linear) models are used to explain spatially distributed data, with the residual variance being accounted for by a spatially varying random function. Another important connection with the classical statistics literature is with the notion of {\it emulation} \cite{kennedy2001bayesian}. Emulation constructs a statistical model of the output of a complex computational model by interpolating sparse model results with a Gaussian Process. Our approach can be viewed as a {\it partial emulation}, whereby we are interested in retaining mechanistic detail for some aspects of the process, and emulate statistically the residual variability. In this light, our work represents a novel approach to bridge the divide between the model-reduction techniques of formal computational modelling and the statistical approximations to complex models.

{\small
       \bibliographystyle{plain}

}

\clearpage

\appendix

\section{Appendix}

All the code that replicate these analysis is available at  the corresponding author's webpage, and hosted on {\sf Github} (repository {\sf  GP-correction-maps}).

\subsection{Further details on the examples} \label{sec:app_ex}

The two models from \S \ref{sec:examp1} correspond to these systems of  differential equations
\begin{center}
\begin{tabular}{cc}
${\bM} {:=} 
\begin{cases} 
\deriv{\E{}}  =  -k_1 \E{} \cdot \SU{} + k_{-1} \ES{} + k_2 \ES{}\\
\deriv{\SU{}}  =  -k_1 \E{}  \cdot\SU{} + k_{-1} \ES{} \\
\deriv{\ES{}}  =  k_1 \E{} \cdot  \SU{} - k_{-1} \ES{} - k_2 \ES{}\\
\deriv{\PR{}}  =   + k_2 \ES{}\\
\end{cases}
\qquad$ &
${\bm} {:=}
\begin{cases} 
\deriv{\E{}}  =  0\\
\deriv{\SU{}}  =  - V_M\SU{} / (K_M + \SU{})  \\
\deriv{\ES{}}  = 0\\
\deriv{\PR{}}  =   + V_M\SU{} / (K_M + \SU{}) \\
\end{cases}
$\end{tabular}
\end{center}
which we solved in \MATLAB{} with the {\sf ode45} routine with all parameters ({\sf InitialStep},  {\sf MaxStep}, {\sf RelTol} and {\sf AbsTol}) set to $0.01$. 

 Concerning the Protein Translation Network (\PTN{}) in \S \ref{sec:examp2},  the set of reactions and their propensity functions that we can use to derive a Continuous Time Markov Chain model of the network are the following. Here $\xx$ denotes a generic state of the system and, for instance,  $\xx_\RNA{}$ the number of \RNA{} copies in $\xx$.
 
\begin{center}
\begin{tabular}{p{3.0cm}p{5.0cm}l}
{\em event} & {\em reaction} & {\em propensity}\\\hline
{\sf activation}& $\GEI \to \GEA$ &$ a_1(\xx) =  {k_{\text{\sf on}}} \cdot \xx_\GEI$\\
{\sf inactivation} & $\GEA \to \GEI$ &$ a_2(\xx) = {k_{\text{\sf off}}\cdot\xx_\PR{}}$\\
{\sf transcription} & $\GEA \to \GEA+\RNA$ &$a_3(\xx) = \alpha\cdot\xx_\GEA{}$\\
{\sf \RNA{} decay} & $\RNA \to \varnothing$ &$a_4(\xx) = {\delta_{\text{\sf RNA}}}\cdot\xx_\RNA{}$\\
{\sf translation} & $\RNA \to \RNA+\PR$ &$a_5(\xx) = \beta\cdot\xx_\RNA{}$\\
{\sf \PR{} decay} & $\PR \to \varnothing$ &$a_4(\xx) = {\delta_{\text{\sf P}}}\cdot\xx_\PR{}$
\end{tabular}
\end{center}
The reduced \PTN{} model is a special of this reactions set where    {\sf transcription} and {\sf \RNA{} decay} are omitted. In this case we used {\sf StochPy} to simulate the models and generate the input data per regression -- see \url{http://stochpy.sourceforge.net/}; data sampling exploits {\sf python} parallelism to reduce execution times.

For regression, we used the {\sf Gaussian Processes for Machine Learning} toolbox for fixed-variance regression, see \url{http://www.gaussianprocess.org/gpml/code/matlab/doc/} and a custom implementation of the other forms of regression.

\begin{figure}[t]
\center
\includegraphics[width=.8\textwidth]{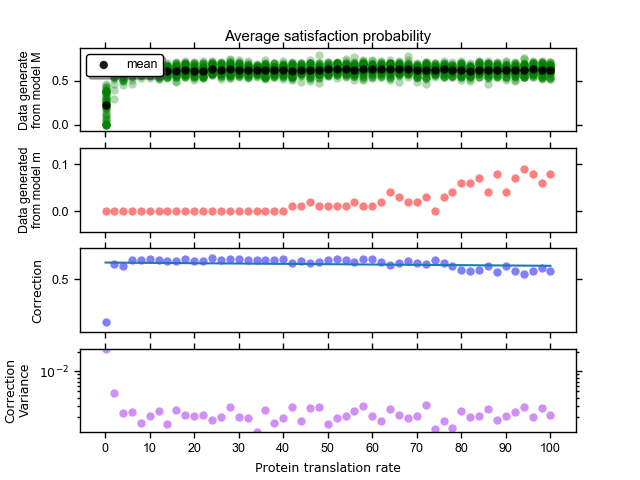}
\caption{Data generated to compute the satisfaction probability of the linear logic formula $\tstat_2$ in eq. (\ref{eq:stat-PTN}). For each model $100$ independent simulations are used to estimate the expectation of the probability. The regression input space is the same used to compute $\tstat_1$, but the models are simulated with just one inactive gene in the initial state. The heteroscedastic variance in the regression is computed as the variance of the correction of the expected satisfaction probability (point-wise $\sigma$-estimator); the fixed-variance regression is computed by estimating the variance from the data (empirical $\overline{\sigma}$-estimator).}
\label{fig:TR-prob-data}
\end{figure}

\subsection{Proofs} \label{app:proofs}

Proof of Theorem 1.
\begin{proof}
Both the empiricals and nested estimator rely on an {\em unbiased} estimator of the mean/variance, which means that if $k\to \infty$, i.e., we sample all possible values for the free variables, we would have a true model of  $\overline{y}$/$\sigma$. This means that, for each sampled value from $\Theta$, even the simplest $\overline{\sigma}$-estimator would be  equivalent, in expectation, to the marginalization of the free variables.  This is enough, combined with  properties of Gaussian Processes regression (i.e., the convergence to the true model with infinite training points), to state that the overall approach leads to an unbiased estimator of the correction map. 
\end{proof}

\end{document}